%% file: main.tex
\documentclass[10pt,twocolumn,letterpaper]{article}

\usepackage{cvpr}              % To produce the CAMERA-READY version
\usepackage{graphicx}
\usepackage{amsmath}
\usepackage{amssymb}
\usepackage{booktabs}
\usepackage{cite}

\usepackage[pagebackref,breaklinks,colorlinks]{hyperref}

\usepackage[capitalize]{cleveref}
\crefname{section}{Sec.}{Secs.}
\Crefname{section}{Section}{Sections}
\Crefname{table}{Table}{Tables}
\crefname{table}{Tab.}{Tabs.}

\def\cvprPaperID{*****} % *** Enter the CVPR Paper ID here
\def\confName{CVPR}
\def\confYear{2022}

\begin{document}

%\title{\LaTeX\ Author Guidelines for \confName~Proceedings}
\title{Multi-modal Emotion Estimation for in-the-wild Videos}
%

\iffalse
\author{First Author\\
Institution1\\
Institution1 address\\
{\tt\small firstauthor@i1.org}
\and
Second Author\\
Institution2\\
First line of institution2 address\\
{\tt\small secondauthor@i2.org}
}
\fi

\author{
Liyu Meng\footnotemark[1], Yuchen Liu\footnotemark[2], Xiaolong Liu\footnotemark[1], Zhaopei Huang\footnotemark[2], \\
Yuan Cheng\footnotemark[3], Meng Wang\footnotemark[3], Chuanhe Liu\footnotemark[1], and  Qin Jin\footnotemark[2] \\
 \footnotemark[1] \small{Beijing Seek Truth Data Technology Co.,Ltd.} \\
 \footnotemark[2] \small{School of Information, Renmin University of China} \\
 \footnotemark[3] \small{Ant Group}
}
%Beijing Seek Truth Data Technology Co.,Ltd.
%School of Information, Renmin University of China

\maketitle

\input{sub_modules/Abstract}
\input{sub_modules/Introduction}
\input{sub_modules/Related_Works}
\input{sub_modules/Method}

\input{sub_modules/Experiments}
\input{sub_modules/Conclusion}

\bibliographystyle{bibstyle}
\bibliography{main}

\end{document}

%% file: sub_modules/Abstract.tex
\begin{abstract}
   In this paper, we briefly introduce our submission to the Valence-Arousal Estimation Challenge of the 3rd Affective Behavior Analysis in-the-wild (ABAW) competition. Our method utilizes the multi-modal information, i.e., the visual and audio information, and employs a temporal encoder to model the temporal context in the videos. 
   Besides, a smooth processor is applied to get more reasonable predictions, and a model ensemble strategy is used to improve the performance of our proposed method.
   The experiment results show that our method achieves $65.55\%$ ccc for valence and $70.88\%$ ccc for arousal on the validation set of the Aff-Wild2 dataset, which prove the effectiveness of our proposed method. 
\end{abstract}

%% file: sub_modules/Introduction.tex
\section{Introduction}
As a crucial part of human-computer interaction, affective computing can be widely used in medical, market analysis, social and other interaction scenarios, and it has extremely indispensable theoretical significance and practical application value to realize humanized communication between human and machine. However, emotions usually arise in response to either an internal or external event which has a positive or negative meaning to an individual\cite{salovey1990emotional}. When recognizing emotions, subtle differences in emotional expressions can also produce ambiguity or uncertainty in emotion perception. Fortunately, with the continuous research in psychology and the rapid development of deep learning, affective computing is gaining more and more attention, for example, Aff-wild\cite{kollias2017recognition, kollias2019deep, zafeiriou2017aff},  Aff-wild2\cite{kollias2022abaw, kollias2021analysing, kollias2020analysing, kollias2021distribution, kollias2021affect, kollias2019expression, kollias2019face, kollias2019deep, zafeiriou2017aff} have provided us with a large-scale dataset of hard labels, driving the development of affective computing. 

In the field of single modality emotion recognition, unimodal information is susceptible to various noises and can hardly reflect the complete emotional state. Multimodal emotion recognition can effectively utilize the information contained in multiple modal recognition, capture the complementary information between modalities, and thus improve the recognition ability and generalization ability of the model. 

In this paper, we adopt Multimodal Representation to perform Multi-modal Fusion of audio features and visual features, and map multimodal information into a unified multimodal vector space. Then in the encoder section we propose to estimate facial empression employ a temporal encoder to model the temporal context in the video. Specifically, two kinds of structures are utilized as the temporal encoder, LSTM\cite{sak2014long} and Transformer\cite{vaswani2017attention}. 
After the temporal encoder, classification is performed through a fully connected layer, and the prediction results are output after the post-processing. Our approach effectively unifies visual and audio embedding into the temporal model and combines transformer and LSTM to design an efficient emotion recognition network to improve the evaluation accuracy of valence and arousal.

%% file: sub_modules/Related_Works.tex
\section{Related Works}
With the 3rd ABAW Competition, researchers from around the world have implemented their latest techniques on the Aff-wild2\cite{kollias2022abaw, kollias2021analysing, kollias2020analysing, kollias2021distribution, kollias2021affect, kollias2019expression, kollias2019face, kollias2019deep, zafeiriou2017aff} dataset. We briefly review some studies on the this competition , there are many approaches on deep learning for face expression analysis. For example: feature fusion , attention mechanisms and iterative self-distillation, which provide us with good inspiration.

On feature fusion, Mollahosseini et al. \cite{vielzeuf2017temporal}  proposed a temporal fusion approach to focus on the problem of multimodal features and temporal features. Multimodal representation learning, which aims to narrow the heterogeneity gap among different modalities, plays an indispensable role in the utilization of ubiquitous multimodal data\cite{8715409}. 
%For instance, \cite{liu2018multi}\cite{liu2020group} use both audio and video channel feature to analysis emotion in video clips, achieved incredible effects.

About the encoding, CAER-Net\cite{lee2019context} proposed an attention-based mechanism that can be used to assist in emotion recognition using context features. Based on attention mechanism, the role of the context part is more interpretable, however, this may lead to a certain degree of feature redundancy. Farzaneh et al. \cite{farzaneh2020discriminant} proposed Discriminant Distribution-Agnostic loss (DDA loss) to regulate the distribution of deep features. With the help of DDA loss, features rich in semantic information for facial expression recognition can increase inter-class seperation and decrease intra-class variations, despite training on unbalanced datasets.

In this work, our contribution is a multi-modal fusion of audio and visual features, using two LSTM and Transformer Encoder structures to obtain a temporal encoder for emotion recognition. Besides, we demonstrate the effectiveness of this approach on valence and arousal recognition on the Aff-wild2 dataset.

%% file: sub_modules/Method.tex
\section{Method}
In this section, we introduce our method for the Valence-Arousal Estimation Challenge in the 3rd ABAW Competition.

\subsection{Overview}
Given a video with sound $X$, it can be divided into the visual data $X^{vis}$ and the audio data $X^{aud}$, where $X^v$ can be illustrated as a sequence of image frames $\{F_{1}, F_{2}, ..., F_{n}\}$, and $n$ denotes the number of image frames contained by $X$. In valence-arousal estimation task, each frame in $X$ is annotated with a sentiment label $Y$ consists of a valence label $Y^{v}$ and an arousal label $Y^{a}$. The task is to predict the sentiment
label corresponding to each frame in the video. 

%The overall pipeline is illustrated in Fig. (a figure), which consists of five components. 
The overall pipeline consists of five components. 
First, all videos are processed to get independent image frames with facial expression on it. Secondly, we extract the visual and audio features corresponding to each frame in the videos. Thirdly, the features are fed into a temporal encoder to model the temporal context in the video, and a full-connected layer is employed to predict the sentiment labels. Finally, the predicted labels pass through some post processors to get the final predictions.

\subsection{Pre-processing}
The videos are first splitted into image frames, and a face detector is applied to get the face bounding box and facial landmarks in each image. Then, the face in each image is cropped out according to the bounding box, and these cropped images are aligned based on the facial landmarks. Here we simply utilize the cropped and aligned facial images provided by the ABAW competition officials. 

In addition, some of the frames don't contain valid faces because the faces in them are not detected or there is no face in them. As for an invalid frame, we find the nearest valid frame around it, and fill it with this valid frame.

\subsection{Feature Extraction}

We use three pre-trained models to extract the visual features. They are the IResNet100-based\cite{duta2021improved} facial expression model, the DenseNet-based\cite{iandola2014densenet} facial expression model and the IResNet100-based Facial Action Units (FAUs) model. We also extract four audio features to increase the performance of our work, they are the eGeMAPS\cite{eyben2015geneva}, the ComParE 2016\cite{schuller2016interspeech}, the VGGish\cite{hershey2017cnn} and the wav2vec2.0\cite{baevski2020wav2vec}.

\textbf{Visual Features:} 
One of the facial expression feature vector is extracted based on independent facial images with a CNN extractor. Specifically, a DenseNet model pre-trained on the FER+ and the AffectNet datasets is employed to extract the visual features. The dimension of the DenseNet-based visual features is 342.

The other facial expression model is pre-trained on the FER+\cite{BarsoumICMI2016}, the RAF-DB \cite{li2017reliable}\cite{li2019reliable} and the AffectNet\cite{mollahosseini2017affectnet} datasets. The network of the model is IResNet100. Faces are aligned by the five face keypoints, we then resize the face into 112x112 as the input of the network. We use the last fully connection layer as the visual feature, the dimension of the vector is 512.

We also use the IResNet100-based model to train the Facial Action Units (FAUs) classifier, whose output is 512-dimensional visual feature vector. 
In particular, the model is pre-trained on the Glint360K\cite{an2021partial} dataset for face recognition at first. And then we fine-tuned the model on the FAU dataset. 

\textbf{Audio Features:} 
The audio features are composed of manually designed low level descriptors (LLDs) and more semantically informative features extracted by deep learning methods. The LLDs contain the eGeMAPS and the ComParE 2016, both of them are extracted by the the openSmile. In our elaborate scheme, the high-level features are extracted by the VGGish and the wav2vec2.0. 
The wav2vec2.0 is a self-supervised model, and the model we use is pre-trained and fine-tuned on 960 hours of the Librispeech\cite{panayotov2015librispeech}. In order to align with the image in the video, the hop size of wav2vec for extracting features is 20ms. The average of the features from the two closest frames of the video image is used as the extracted feature. The VGGish is mainly used for speech classification, and it is pretrained on a large youtubue dataset (Audioset\cite{gemmeke2017audio}). The embedding output of this model is 128 dimensions.

\subsection{Architectures}
Due to the limitation of GPU memory, we split the videos into segments at first. Given the segment length $l$ and stride $p$, a video with $n$ frames would be split into $[n/p]+1$ segments, where the $i$-th segment contains frames $\{F_{(i-1)*p+1}, ..., F_{(i-1)*p+l}\}$. 
\subsubsection{Multi-modal Fusion}
Given the visual features $f^{v}_{i}$ and audio features $f^{a}_{i}$ corresponding to the $i$-th segment, they are first concatnated and fed into a full-connected layer to get the multi-modal features $f^{m}_{i}$. It can be formulated as follows:

\begin{small}
\begin{equation}
    \centering
    f^{m}_{i} = W_{f}[f^{v}_{i};f^{a}_{i}] + b_{f}
\end{equation}
\end{small}
where $W_{f}$ and $b_{f}$ are learnable parameters. 

\subsubsection{Temporal Encoder}
With the multi-modal features, We employ a temporal encoder to model the temporal context in the video. Specifically, two kinds of structures are utilized as the temporal encoder, including LSTM and Transformer Encoder.

\textbf{LSTM}
We employ a Long Short-Term Memory Network (LSTM) to model the sequential dependencies in the video. For the $i$-th video segment $s_{i}$, the multi-modal features $f^{m}_{i}$ are directly fed into the LSTM. In addition, the last hidden states of the previous segment $s_{i-1}$ are also fed into the LSTM to encode the context between two adjacent segments. It can be formulated as follows:
\begin{small}
\begin{equation}
    \centering
    g_{i}, h_{i} = \text{LSTM}(f^{m}_{i}, h_{i-1})
\end{equation}
\end{small}
where $h_{i}$ denotes the hidden states at the end of $s_{i}$. $h_{0}$ is initialized to be zeros. In order to ensure that the last frame of $s_{i-1}$ and the first frame of segment $s_{i}$ are consecutive frames, there is no overlap between two adjacent segments when LSTM is used as the temporal encoder. In another word, the stride $p$ is the same as the segment length $l$.

\textbf{Transformer Encoder}
We utilize a transformer encoder to model the temporal information in the video segment as well, which can be formulated as follows:

\begin{small}
\begin{equation}
    \centering
    g_{i} = \text{TRMEncoder}(f^{m}_{i})
\end{equation}
\end{small}

Unlike LSTM, the transformer encoder just models the context in a single segment and ignores the dependencies of frames between segments. In order to cover context of different frames, there can be overlaps between consecutive segments, which means  $p \leq l$.

\subsubsection{Regression}
After the temporal encoder, the features $g_{i}$ are finally fed into full-connected layers for regression, which can be formulated as follows:

\begin{small}
\begin{equation}
    \centering
    \hat{y}_{i} = W_{p}g_{i} + b_{p}
\end{equation}
\end{small}
where $W_{p}$ and $b_{p}$ learnable parameters, $\hat{y}_{i} \in \mathbb{R}^{l \times 2}$ are the predictions of the valence and arousal labels of $s_{i}$.

\subsubsection{Training Objects}
We use the Concordance Correlation Coefficient (CCC) between the predictions and the ground truth labels as the loss function, which can be denoted as follows:

\begin{small}
\begin{equation}
    \centering
    L = \sum_{c \in \{v, a\}} (1 - CCC(\hat{y}^{c}, y^{c}))
\end{equation}
\end{small}
where $\hat{y}^{v}, \hat{y}^{a}, y^{v}, y^{a}$ denotes the predictions and the ground truth labels of valence and arousal in a batch respectively. 

\subsection{Post-processing}
In the testing stage, we apply some additional post processors to the predictions. First, some of the predictions may exceed the range of $[-1, 1]$, and we simply cut these values to $-1$ or $1$.

Secondly, since the sentiment of individuals varies continuously over time, the value of valence and arousal also varies smoothly over time. Thus, we apply a smooth function to the predictions to make them smooth in time. Specifically, given the original prediction of the $j$-th frame $\hat{y}_{j}$, the final prediction $\tilde{y}_{j}$ is set as the average prediction value of a window with $w$ frames centered on the $j$-th frame, i.e.,  $\{\hat{y}_{j-[w/2]}, ..., \hat{y}_{j+[w/2]}\}$.

%% file: sub_modules/Experiments.tex
\begin{table*}[]
\begin{center}
  \caption{The performance of our method on the validation dataset.}
  \label{tab:main}
\begin{tabular}{c|c|c|c|c}
\hline
Model          & Visual Features & Audio Features                 & Valence         & Arousal         \\ \hline
LSTM           & DenseNet        & wav2vec                        & 0.5544          & 0.6531          \\
TRM-v1 & DenseNet        & wav2vec, ComParE               & \textbf{0.6050} & 0.6416          \\
TRM-v2 & ires100,fau     & wav2vec,VGGish,ComParE,eGeMAPS & 0.5883          & \textbf{0.6689} \\ \hline
\end{tabular}
\end{center}
\end{table*}

\begin{table*}[]
\begin{center}
  \caption{Ablation study of features on the validation dataset.}
  \label{tab:ablation}
\begin{tabular}{c|c|c|c|c}
\hline
Model          & Visual Features                    & Audio Features                    & Valence     & Arousal      \\ \hline
TRM-v1 & DenseNet                           & None                              & 0.5290          & 0.5969          \\
TRM-v1 & DenseNet                           & wav2vec                           & 0.5596          & 0.6460          \\
TRM-v1 & DenseNet                           & wav2vec, VGGish                   & 0.5663          & 0.6464               \\
TRM-v1 & DenseNet                           & wav2vec, ComParE                  & \textbf{0.6050} & 0.6416          \\ \hline
TRM-v2 & ires100     & wav2vec, VGGish, ComParE, eGeMAPS & 0.5055          & 0.6166          \\
TRM-v2 & fau         & wav2vec, VGGish, ComParE, eGeMAPS & 0.5707          & 0.6168          \\
TRM-v2 & ires100,fau & wav2vec                           & 0.5357          & 0.6412          \\
TRM-v2 & ires100,fau & wav2vec,VGGish                    & 0.5843          & 0.6614          \\
TRM-v2 & ires100,fau & wav2vec,VGGish,ComParE,eGeMAPS    & 0.5883          & \textbf{0.6689} \\ \hline
\end{tabular}
\end{center}
\end{table*}

\section{Experiments}
\subsection{Dataset}
The third ABAW competition aims to automate affective analysis and includes four challenges: i) uni-task Valence-Arousal Estimation, ii) uni-task Expression Classification, iii) uni-task Action Unit Detection, and iv) MultiTask-Learning. All challenges are based on a common benchmark database, Aff-Wild2, a large-scale field database and the first to be annotated according to valence arousal, expression and action units. 
the Aff-Wild2 database extends the Aff-Wild, with more videos and annotations for all behavior tasks.
The Valence-Arousal Estimation Challenge contains 567 videos, have been annotated by four experts using the method proposed in \cite{cowie2000feeltrace}.

For the face expression recognition model, we used FER+, RAF-DB and AffectNet for pre-training. The FER+ dataset is relabeled from the fer2013 \cite{carrier2013fer} dataset. The fer2013 contains: Angry, Disgust, Fear, Happy, Neutral, Sad and Surprise.
The RAF-DB is a large-scale database of facial expressions, which includes about 30,000 images of a wide variety of faces downloaded from the Internet. We use the single-label subset in RAF-DB, including 7 classes of basic emotion.
AffectNet contains over one million facial images, collected from the Internet. Approximately half of the retrieved images (approximately 440,000) were manually annotated for the presence of seven discrete facial expressions (classification model) as well as intensity of value and arousal.

\subsection{Experimental Settings}
The models are trained on Nvidia GeForce GTX 1080 Ti GPUs, 
each with 11GB memory, and with the Adam \cite{kingma2014adam} optimizer. The results reported in the following experiments are based on the average score of 3 random runs.
The model is trained for 30 epochs, the batch size is 16 and the dropout rate is 0.3.
As for the LSTM model, the learning rate is 0.0003, the dimension of multi-modal features and the hidden size are 512, the length of video segments is 100, the number of regression layers is 2 and the hidden size are \{512, 256\} respectively.

As for the transformer encoder model, two sets of hyper-parameters are used, which are called \textbf{TRM-v1} and \textbf{TRM-v2}. 
The hyper-parameters of TRM-v1 and -v2 are shown as follows respectively: the learning rate is \{0.0002, 0.0003\}, the length of video segments is \{250, 250\}, the stride of segments is \{250, 100\}, the dimension of multi-modal features is \{256, 512\}, the numbers of encoder layers is \{4, 4\}, the number of attention heads is \{4, 4\}, the dimension of feed forward layers in the encoder is \{1024, 512\}, the number of regression layers is \{2, 2\} and the hidden size of regression layers are \{256, 256\} for TRM-v1 and \{512, 256\} for TRM-v2. 
As for the smooth fuction in the post processing stage, the size of the smoothing window is 
20 for valence and 50 for arousal.

\subsection{Overall Results}

Table \ref{tab:main} shows the experimental results of our proposed method on the validation set of the Aff-Wild2 dataset. The Concordance Correlation Coefficient (CCC) is used as the evolution metrics for both valence and arousal prediction task. As is shown in the table, our proposed TRM-v1 and -v2 structures achieve the best performance for valence and arousal respectively, and the LSTM structure achieves competitive performance for arousal as well. It prove the effectiveness of each of our proposed structures.

\subsection{Ablation Study}
In this section, we conduct an ablation analysis of different features to compare the contribution of them. Table \ref{tab:ablation} shows the results of the ablation study for our proposed TRM-v1 and -v2. 

As is shown in the table, each of our proposed features has contributed to the performance. As for the audio features, the ComParE and wav2vec make the most contributions for the valence prediction task, while the VGGish and wav2vec maks the most contributions for arousal. As for the visual features, FAU contributes more than IRES100 for valence, and DenseNet contributes more than the combination of FAU and IRES100 for valence, while less for arousal.

\begin{table}[]
\begin{center}
  \caption{The results of each single model and the ensemble of them for the valence prediction task.}
  \label{tab:valence}
\resizebox{\columnwidth}{!}{
\begin{tabular}{c|c|c}
\hline
Model    & Features                                & Valence         \\ \hline
TRM-v1   & DenseNet,wav2vec,ComParE                & 0.6089          \\
TRM-v1   & DenseNet,wav2vec,ComparE,VGGish,eGeMAPs & 0.6113          \\
TRM-v2   & ires100,fau,wav2vec,VGGish              & 0.5833          \\
TRM-v2   & ires100,fau,VGGish,ComParE,eGeMAPS      & 0.5831          \\ \hline
Ensemble &                                         & \textbf{0.6555} \\ \hline
\end{tabular}
}
\end{center}
\end{table}

\begin{table}[]
\begin{center}
  \caption{The results of each single model and the ensemble of them for the arousal prediction task.}
  \label{tab:arousal}
 
\resizebox{\columnwidth}{!}{
\begin{tabular}{c|c|c}
\hline
Model    & Features                                      & Arousal         \\ \hline
LSTM     & DenseNet,wav2vec                            & 0.6591          \\
TRM-v1   & DenseNet,wav2vec                           & 0.6488          \\
TRM-v1   & DenseNet,wav2vec,ComParE                   & 0.6458          \\
TRM-v1   & DenseNet,wav2vec,VGGish,eGeMAPs            & 0.6456          \\
TRM-v2   & ires100,fau,wav2vec,VGGish                 & 0.6628          \\
TRM-v2   & ires100,fau,wav2vec,VGGish,ComParE,eGeMAPS & 0.6604          \\ \hline
Ensemble &                                            & \textbf{0.7088} \\ \hline
\end{tabular}
}
\end{center}
\end{table}

\subsection{Model Ensembles}
In order to further improve the performance of our proposed models, we apply a model ensemble strategy to these models. We train some models with different basic structures, hyper-parameters and combination of features, and get the predictions of them respectively in the testing stage. Then, the average value of the prediction of these models are taken as the final prediction. 

Table \ref{tab:valence} and Table \ref{tab:arousal} show the results of model ensembles for the valence and arousal prediction task respectively. The results indicates that the model ensemble strategy can combine the strengths of different models and achieve significant improvement over them.

%% file: sub_modules/Conclusion.tex
\section{Conclusion}

In this paper, we introduce our method for the Valence-Arousal Estimation Challenge of the 3rd Affective Behavior Analysis in-the-wild (ABAW) competition. 
Our method utilizes the multi-modal information and employs a temporal encoder to model the temporal context in the videos. 
Besides, some post processors are used to improve the performance of our proposed method.
The experiment results show that our method achieves $65.55\%$ ccc for valence and $70.88\%$ ccc for arousal on the validation set of the Aff-Wild2 dataset, which prove the effectiveness of our proposed method.